\documentclass{amia}
\usepackage{lipsum} %

\usepackage{caption}
\usepackage{subcaption}
\usepackage{multirow}
\usepackage{arydshln, comment}
\usepackage[table,xcdraw]{xcolor}
\usepackage{wrapfig,booktabs}
\usepackage{physics,amssymb}
\usepackage[flushleft]{threeparttable}

\newcommand{\revise}[1]{{#1}}

\begin{document}

\title{Split Learning for Distributed Collaborative\\Training of Deep Learning Models in Health Informatics}

\author{Zhuohang Li, MS$^1$, Chao Yan, PhD$^2$, Xinmeng Zhang, BS$^1$, Gharib Gharibi, PhD$^3$,\\Zhijun Yin, PhD$^{1,2}$, Xiaoqian Jiang, PhD$^4$, Bradley A. Malin, PhD$^{1,2}$ }

\institutes{
    $^1$\mbox{Vanderbilt University, Nashville, TN;$^2$Vanderbilt University Medical Center, Nashville, TN;}\\$^3$TripleBlind, Kansas City, MO;$^4$UTHealth, Houston, TX
}

\maketitle

\section*{Abstract}
\vspace{-2mm}

\textit{Deep learning continues to rapidly evolve and is now demonstrating remarkable potential for numerous medical prediction tasks. However, realizing deep learning models that generalize across healthcare organizations is challenging. This is due, in part, to the inherent siloed nature of these organizations and patient privacy requirements. To address this problem, we illustrate how split learning can enable collaborative training of deep learning models across disparate and privately maintained health datasets, while keeping the original records and model parameters private. We introduce a new privacy-preserving distributed learning framework that offers a higher level of privacy compared to conventional federated learning. We use several biomedical imaging and electronic health record (EHR) datasets to show that deep learning models trained via split learning can achieve highly similar performance to their centralized and federated counterparts while greatly improving computational efficiency and reducing privacy risks.}

\vspace{-1mm}
\section*{Introduction}
\vspace{-2mm}
Recent advances in deep learning algorithms have enabled the development of neural networks with promising performance for a variety of healthcare data types that were previously considered challenging with traditional statistical methods, including medical images~\cite{esteva2017dermatologist}, natural languages~\cite{wen2019desiderata}, and structured electronic health records (EHR)~\cite{landi2020deep}.
Despite the remarkable progress that has been achieved, most current deep learning models in the healthcare domain are developed using data from only a single site.
Yet it is evident that no single healthcare organization can collect a sufficient amount of data on diverse populations, as well as variations in organizational practices, to represent the distribution of the general patient population. 
Consequently, models developed from single-site data oftentimes lack sufficient generalizability and may perform poorly when applied to other sites~\cite{cohen2021problems}.
To resolve this problem, one natural solution would be to enable disparate healthcare organizations to collaboratively train a model. However, such collaborations have met numerous obstacles, ranging from concerns over data rights to patient privacy.

Over the past several years, distributed learning has been investigated as a strategy to enable multiple data holders to contribute to the development of deep learning models while protecting the privacy of the underlying raw records.
Specifically, one of the most popular variants of distributed learning is \textit{federated learning} (FL)~\cite{mcmahan2017communication,rieke2020future}, which enables data holders to contribute to the training of a learning model under the orchestration of a central server by exchanging only focused model updates while maintaining private data locally. The ability to maintain the privacy of record-level data has drawn particular research interest from the healthcare community~\cite{kaissis2020secure,dayan2021federated}.
Still, the canonical form of federated learning requires model builders to reveal the details about their local models (i.e., model architecture and model parameters) and such information can be leveraged to make inferences about the privately maintained local data records. 
As a result, in untrustworthy environments, federated learning needs to be jointly implemented in concert with additional mechanisms
to enhance its privacy support. There are various approaches for doing so; some of the popular methods include 1) \textit{differential privacy} (DP)~\cite{abadi2016deep,truex2020ldp} which leverages a randomization mechanism (e.g., additive Laplacian noise) to provide privacy guarantees for individual records for algorithms on aggregate databases,
2) \textit{secure multiparty computation} (SMC)~\cite{constable2015privacy,bonawitz2017practical}
a cryptographic solution to enable a set of parties to compute a joint function on their private data without revealing anything but the prescribed output
and 3) \textit{homomorphic encryption} (HE)~\cite{froelicher2021truly,hardy2017private}, which allows a party to compute certain mathematical operations on ciphertexts without decrypting them.
However, in practice, these additional privacy protection measures often come at the cost of either significantly harming model utility (e.g., predictive performance) or increasing computational complexity, which leads to long runtimes and costly computation in cloud computing environments.

In this paper, we investigate \textit{split learning} (SL) as a new paradigm for multi-institutional collaborative learning of deep neural networks across distributed health data.
Similar to conventional distributed learning, the split learning framework is composed of a server (i.e., the coordinator) and multiple healthcare organizations (i.e., the data contributors)~\cite{gupta2018distributed}.
Its uniqueness is that under the split learning setting, the deep neural network, \revise{also referred to as the global model,} is divided into two sub-models (i.e., the \textit{client model} and the \textit{server model}) according to a specific layer known as the \textit{cut layer}~\cite{vepakomma2018split}. The healthcare organizations and the server hold only their portion of the model, which they do not share with each other.
At each training round, the healthcare organizations only train the first part of a deep neural network and send, what the literature has called, the \textit{smashed data} (i.e., the latent representations of raw data derived from the client model) to the server. The server then completes the rest of the forward propagation and computes the loss function without accessing the clients' raw data. Finally, the training round is concluded with a backward pass, where the server sends back to the clients the computed gradients, which are then used to update the client model. This process is equivalent to a training epoch in the centralized learning setting and will iterate until the global model converges.

Table~\ref{tab:framework} provides a qualitative comparison between different distributed learning frameworks. Notably, as is in the federated learning setting, no sensitive raw data is shared during the split learning training process, which maintains the privacy of patients' data. Moreover, in split learning, neither the server nor the clients have complete knowledge of the global model's architecture and weights. This is a major difference from federated learning, where the server has full access to the client's model. This notion of incompleteness in knowledge about the global model, combined with the minimal information encoded in the smashed data, further reduces the risk of privacy leakage during training.
This is also beneficial to the server in scenarios where the server hopes not to reveal its developed model architecture, which may be considered proprietary.
In addition to privacy benefits, split learning can also alleviate the computational burden on the healthcare organizations' side by offloading part of the training process of the deep neural network to the server (e.g., a data center) which typically has access to more computational power at a cheaper rate.
\revise{
In addition to the qualitative study, we further 1) provide a quantitative analysis of the split learning framework by conducting experiments across three biomedical image datasets (PathMNIST~\cite{kather2019predicting}, OrganAMNIST~\cite{bilic2019liver}, and BloodMNIST~\cite{acevedo2020dataset}) and two EHR datasets (eICU~\cite{pollard2018eicu} and a private dataset from Vanderbilt University Medical Center~\cite{zhang2021predicting}),
2) perform an analysis to
compare the privacy risk under the split learning and federated learning framework,
and 3) investigate the trade-off between privacy, model utility, and client-side model training efficiency in split learning.
Our results suggest that split learning can consistently achieve comparable performance as federated learning while providing enhanced privacy and computational efficiency for the participating health organizations.}

\begin{table}[]
\centering
\caption{A qualitative comparison of different distributed learning schemes.}\label{tab:framework}
\vspace{-2mm}
\resizebox{0.96\linewidth}{!}{
\begin{threeparttable}
\begin{tabular}{
>{\columncolor[HTML]{FFFFFF}}c ||
>{\columncolor[HTML]{E5FFCC}}c c
>{\columncolor[HTML]{FFCCCC}}c |
>{\columncolor[HTML]{E5FFCC}}c |c}
\hline
\cellcolor[HTML]{FFFFFF}                                     & \multicolumn{3}{c|}{\cellcolor[HTML]{FFFFFF}\textbf{Privacy Implications}}                                                                                                                                                                                                                                                                                                  & \cellcolor[HTML]{FFFFFF}                                         & \cellcolor[HTML]{FFFFFF}                                                    \\ \cline{2-4}
\multirow{-2}{*}{\cellcolor[HTML]{FFFFFF}\textbf{Framework}} & \multicolumn{1}{c|}{\cellcolor[HTML]{FFFFFF}\textbf{\begin{tabular}[c]{@{}c@{}}Protection over\\ Raw Data\end{tabular}}} & \multicolumn{1}{c|}{\cellcolor[HTML]{FFFFFF}\textbf{\begin{tabular}[c]{@{}c@{}}Protection over\\ Model Parameters\end{tabular}}} & \cellcolor[HTML]{FFFFFF}\textbf{\begin{tabular}[c]{@{}c@{}}Protection over\\ Model Architecture\end{tabular}} & \multirow{-2}{*}{\cellcolor[HTML]{FFFFFF}\textbf{Model Utility}} & \multirow{-2}{*}{\cellcolor[HTML]{FFFFFF}\textbf{Computational Efficiency}} \\ \hline \hline
\textbf{FL~\cite{mcmahan2017communication}}                                                  & \multicolumn{1}{c|}{\cellcolor[HTML]{E5FFCC}Yes}                                                                         & \multicolumn{1}{c|}{\cellcolor[HTML]{FFCCCC}No}                                                                                  & No                                                                                                            & High                                                             & \cellcolor[HTML]{FFFC9E}Moderate                                            \\ \hline
\textbf{FL+DP~\cite{truex2020ldp}}                                               & \multicolumn{1}{c|}{\cellcolor[HTML]{E5FFCC}Yes}                                                                         & \multicolumn{1}{c|}{\cellcolor[HTML]{FFFC9E}Variable$^*$}                                                                                 & No                                                                                                            & \cellcolor[HTML]{FFFC9E}Variable$^*$                                      & \cellcolor[HTML]{FFFC9E}Moderate                                            \\ \hline
\textbf{FL+SMC~\cite{bonawitz2017practical}}   
& \multicolumn{1}{c|}{\cellcolor[HTML]{E5FFCC}Yes}                                                                         & \multicolumn{1}{c|}{\cellcolor[HTML]{FFCCCC}No}                                                                                  & No                                                                                                            & High                                                             & \cellcolor[HTML]{FFCCCC}Low                                                 \\ \hline
\textbf{FL+SMC+HE~\cite{hardy2017private}}                                              & \multicolumn{1}{c|}{\cellcolor[HTML]{E5FFCC}Yes}                                                                         & \multicolumn{1}{c|}{\cellcolor[HTML]{E5FFCC}Yes}                                                                                  & No                                                                                                            & High                                                             & \cellcolor[HTML]{FFCCCC}Low                                                 \\ \hline
\textbf{SL~\cite{vepakomma2018split}}                                                  & \multicolumn{1}{c|}{\cellcolor[HTML]{E5FFCC}Yes}                                                                         & \multicolumn{1}{c|}{\cellcolor[HTML]{FFFC9E}Partial$^\dag$}                                                                             & \cellcolor[HTML]{FFFC9E}Partial$^\dag$                                                                               & High                                                             & \cellcolor[HTML]{E5FFCC}High                                                \\ \hline
\end{tabular}
\begin{tablenotes}
  \small
  \item $^*$ Depends on the choice of privacy parameters and it's typically a trade-off between privacy and utility.
  \item $^\dag$ Associated with the relative location of the cut layer: only the shallow layers of the model (up to the cut layer) are protected from the server.
\end{tablenotes}
\end{threeparttable}
}
\vspace{-2mm}
\end{table}

\vspace{-1mm}
\section*{Method}
\vspace{-1mm}

\begin{figure}[t]
  \centering
  \begin{subfigure}{0.42\linewidth}
    \centering
    \includegraphics[width=\columnwidth]{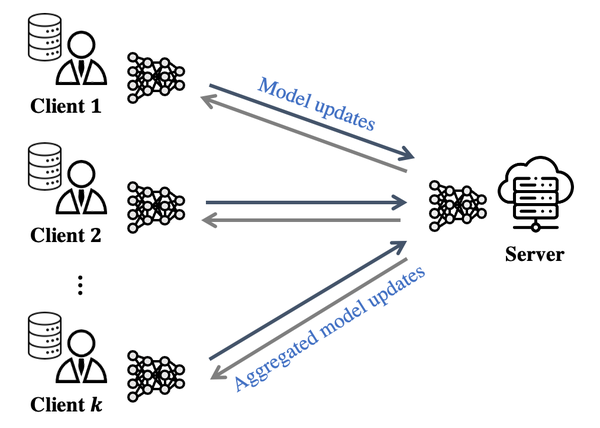}
    \caption{Federated learning}
    \label{subfig:FL}
  \end{subfigure}
  \hspace{2mm}
  \begin{subfigure}{0.42\linewidth}
    \centering
    \includegraphics[width=\columnwidth]{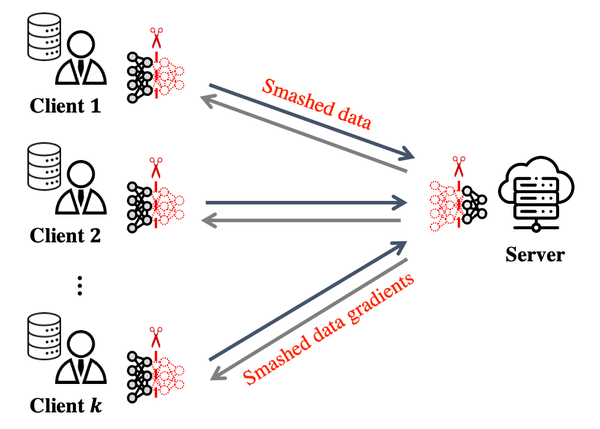}
    \caption{Split learning}
    \label{subfig:SL}
  \end{subfigure}

  \vspace{2mm}
  \caption{Illustration of the federated learning and split learning frameworks.}
  \vspace{1mm}
\end{figure}

\paragraph{\textit{Federated Learning.}}
Federated learning can be conducted among either a small group of organizations (cross-silo) or a large population of mobile devices (cross-device).
In this paper, we focus on the cross-silo setting, since it is more prevalent
in the healthcare domain.
As depicted in Figure\ref{subfig:FL}, we assume that there are $k$ clients who participate in the model training. Each client holds $n_i$ data samples ($i \in \{1, 2, ..., k\}$). $n= \sum_i^k n_i$ is the total data size. The objective of the collaboration is to train a neural network $F_{\vb{w}}$ parameterized by weights $\vb{w}$. At the beginning of the model training, each client initializes its local model $\vb{w}^c_i$ \revise{in parallel}. For each training epoch, each client calculates the loss $\mathcal{L}\big(F_{\vb{w}^c_i}(\vb{X}_i), \vb{y}_i\big)$ using its own data $\vb{X}_i$ with labels $\vb{y}_i$ in parallel. A local model update is computed through gradient descent and shared with the server: $\vb{w}^c_i \leftarrow \vb{w}^c_i - \eta\cdot\frac{\partial \mathcal{L}}{\partial \vb{w}^c_i}$. Finally, at the end of each training epoch, the server updates the global model weights by computing a weighted average of the local models, a process known as federated averaging~\cite{mcmahan2017communication}: $\vb{w} \leftarrow \sum_i^k\frac{n_i}{n}\cdot \vb{w}^c_i$. This process is repeated until the global model converges.

\paragraph{\textit{Split Learning.}}
As shown in Figure~\ref{subfig:SL}, in the split learning setting, a neural network $F$ is partitioned into two separate sub-models. The first is the client model $h_{\vb{w}^c}$, which takes the raw data $\vb{X}$ and outputs latent representations of the data known as the \textit{smashed data}. The second is the server model $f_{\vb{w}^s}$, which makes predictions based on the smashed data, i.e., $F(\vb{X}) = (f_{\vb{w}^s}\circ h_{\vb{w}^c})(\vb{X})$. The clients and the server only have access to their own part of the model and cannot access the other part.

\revise{Each training step of a neural network can be described as a \textit{forward pass} where the loss function is computed and a \textit{backward pass} where the model parameters are updated by back-propagating the error through gradients.}
In the canonical version of split learning, at each forward pass, the client computes the smashed data $h_{\vb{w}^c}(\vb{x})$ and then shares the smashed data along with its label to the server. The server makes a prediction of the smashed data using the server model to compute the loss $\mathcal{L}\Big(f_{\vb{w}^s}\big(h_{\vb{w}^c}(\vb{X})\big), \vb{y}\Big)$. In the backward pass, the server first updates its model according to $\vb{w}^s \leftarrow \vb{w}^s - \eta\cdot \frac{\partial \mathcal{L}}{\partial \vb{w}^s}$ and sends the gradients of the smashed data $\frac{\partial \mathcal{L}}{\partial h_{\vb{w}^c}(\vb{X})}$ to the client. The client then computes the gradients of the client model $\frac{\partial \mathcal{L}}{\partial \vb{w}^c}$ and updates the model parameters accordingly i.e., $\vb{w}^c \leftarrow \vb{w}^c - \eta\cdot \frac{\partial \mathcal{L}}{\partial \vb{w}^c}$, where $\eta$ is the learning rate.
Finally, in the multiple clients' scenario, the client shares its model parameters with the next client where this process repeats.
Note that the raw data is never shared in the process and the server only sees a compact representation of the client's private data (the smashed data). Moreover, the computational cost on the clients' end is greatly reduced since the client is only responsible for the computation of the first part of the model.
\revise{It should be recognized that there are variants of split learning that do not require sharing labels or allow learning on vertically partitioned data~\cite{vepakomma2018split}.}

\paragraph{\textit{Speeding up Split Learning with Federation.}}
The canonical split learning framework processes each client in a sequential order, which can become very time-consuming if either the number of clients or the amount of data maintained by each client is large. Inspired by federated learning, the split learning framework can be modified to enable clients to compute their updates in a parallel manner and thereby accelerate the learning process~\cite{thapa2022splitfed}.
Specifically, at each round of training, $k$ clients first compute their smashed data in parallel $h_{\vb{w}^c_i}(\vb{X}_i), i \in \{1, 2, ..., k\}$ and send the results to the server. Next, the server computes the loss $\mathcal{L}_i\Big(f_{\vb{w}^s_i}\big(h_{\vb{w}^c_i}(\vb{X}_i)\big), \vb{y}_i\Big)$ and the gradients $\frac{\partial \mathcal{L}_i}{\partial \vb{w}^c_i}$ and sends back the gradients to each individual clients. The server then updates its model by averaging the gradients from all clients: $\vb{w}^s \leftarrow \vb{w}^s - \eta \sum_i^k \frac{n_i}{n}\cdot \frac{\partial \mathcal{L}_i}{\partial \vb{w}^s_i}$. Similarly,
the clients update their models via federated averaging, i.e., $\vb{w}^c \leftarrow \vb{w}^c - \eta \sum_i^k \frac{n_i}{n}\cdot\frac{\partial \mathcal{L}_i}{\partial \vb{w}^c_i}$, with the help of a separate server referred to as the federated server.
By allowing parallel computing on both the clients' and the server's side, this variant of split learning (named SplitFedv1 in~\cite{thapa2022splitfed}) can significantly speed up the training process.
However, the drawback of this approach is that to protect the client's data privacy, it is required that the federated server is a trustworthy
third party that does not collude with the main server.
In addition, the federation process may negatively affect the utility of the converged model.
Nevertheless, a low computation latency is often desired for deploying distributed learning solutions to large-scale real-world applications. Thus, we use this variant of split learning as the default in our experiments.

\section*{Evaluation}

\subsection*{\textit{Experimental Setup}}

\paragraph{\textit{Datasets.}}
\begin{wraptable}{r}{0.5\linewidth}
\vspace{-10mm}
\begin{center}
\caption{A summary of the datasets used in this study.}\label{tab:dataset}
\resizebox{\linewidth}{!}{
\begin{tabular}{c|c|c|c}
\hline
\textbf{Dataset}	            &\textbf{Data Type}  	            &\textbf{Task}                          &\textbf{Label Distribution} \\
\hline \hline
PathMNIST           &Colon pathology            &Multi-class    &    \begin{minipage}{.22\textwidth} \centering \includegraphics[width=\textwidth]{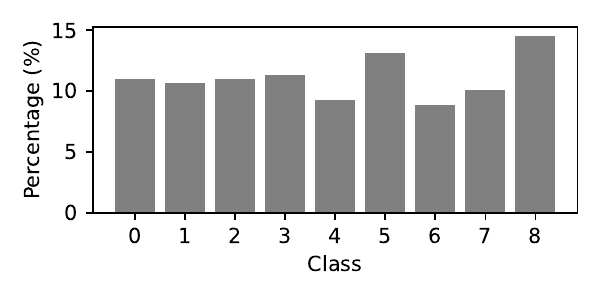} \end{minipage}             \\
\hline 
OrganAMNIST         &Abdominal CT               &Multi-class    &\begin{minipage}{.22\textwidth} \centering \includegraphics[width=\textwidth]{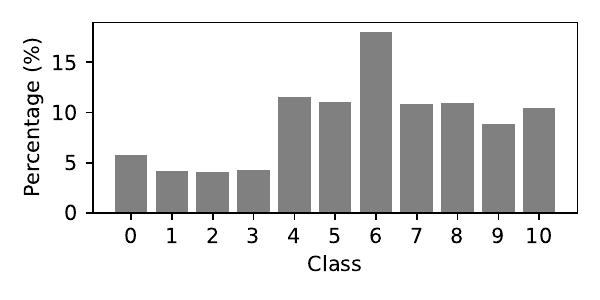} \end{minipage}             \\
\hline
BloodMNIST          &Blood cell microscope      &Multi-class    &\begin{minipage}{.22\textwidth} \centering \includegraphics[width=\textwidth]{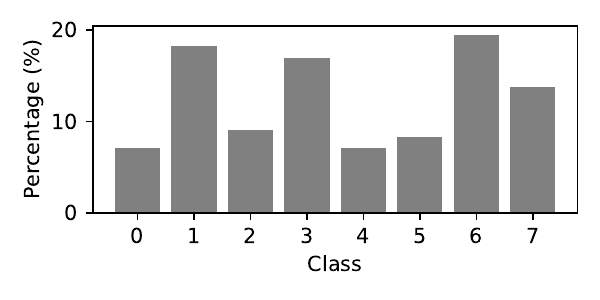} \end{minipage}             \\
\hline
eICU                &Electronic health record   &Binary-class         &\begin{minipage}{.22\textwidth} \centering \includegraphics[width=\textwidth]{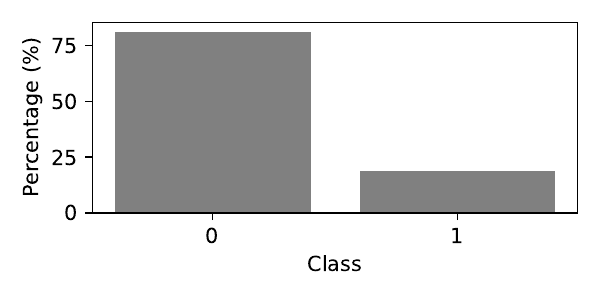} \end{minipage}    \\
\hline
VUMC                &Electronic health record   &Binary-class         &\begin{minipage}{.22\textwidth} \centering \includegraphics[width=\textwidth]{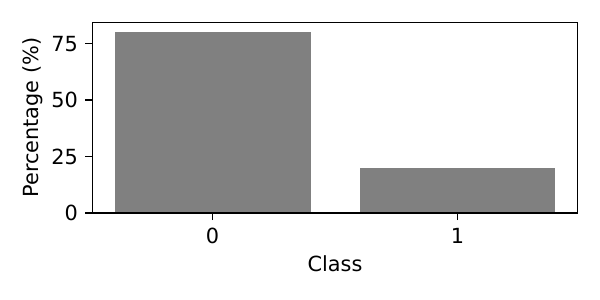} \end{minipage}    \\
\hline
\end{tabular}
}
\vspace{-4mm}
\end{center}
\end{wraptable}
We utilize five datasets to support evaluations in two types of settings - biomedical image classification and clinical concept predictions from structured EHR data.
Specfically, we apply the following three datasets from the MedMNIST~\cite{yang2023medmnist} benchmark for \textit{biomedical image classification} tasks:
(1) PathMNIST~\cite{kather2019predicting}: a colon pathology dataset containing $100,000$ non-overlapping image patches from hematoxylin \& eosin stained histological images for classifying $9$ types of tissues. We follow the recommended train/test split in our experiment, resulting in a total number of $89,996$ images for training. An additional $7,180$ image patches collected from a different clinical center are reserved for testing;
(2) OrganAMNIST~\cite{bilic2019liver}: a 2D image dataset cropped from the axial view of 3D computed tomography (CT) images from Liver Tumor Segmentation Benchmark~\cite{bilic2019liver} for performing classification of $11$ body organs. The dataset is partitioned into two disjoint sets, with $34,581$ images for training and $17,778$ for testing;
and (3) BloodMNIST~\cite{acevedo2020dataset}: a peripheral
blood cell images dataset containing individual normal cells organized into $8$ classes.
The dataset is partitioned into
training and testing sets, each with $11,959$ and $3,421$ images, respectively.
For the \textit{EHR prediction} tasks, we use the following two datasets:
(1) eICU~\cite{pollard2018eicu}: a public EHR dataset containing more than $140,000$ patients' hospital visit records. The task is to predict the risk of being readmitted to the ICU in the next $15$ days (i.e., binary classification)
given the clinical activities during the current ICU stay. Following previous studies~\cite{choi2020learning,yang2023manydg}, we consider five types of events as features, including diagnosis, lab test, medication, physical exam, and treatment.
We use non-overlapping training and testing sets containing data from $30,000$ and $10,000$ patients, respectively;
and (2) VUMC~\cite{zhang2021predicting}: a private EHR dataset collected from Vanderbilt University Medical Center containing all adult inpatient visits in 2019. The task is to predict whether the patient will be discharged the next day to home or other care facilities. Visits shorter than 24h
or of patients who died during hospitalization are excluded, resulting in a total number of $26,283$ patients with an average age of $52.9$.
Table~\ref{tab:dataset} provides a summary of the description and label distribution for each dataset.

\begin{wraptable}{r}{0.3\linewidth}
\centering
\vspace{-6mm}
\caption{Model architecture for biomedical image classification.}
\label{tab:CNN_structure}
\vspace{-2mm}
\resizebox{\linewidth}{!}{
\begin{tabular}{cccc}
\hline
\textbf{Layer}   & \textbf{Kernel} & \textbf{Stride} & \textbf{Output} \\ \hline \hline
Conv2D & $3\times3$    & $1\times1$    & $16$     \\
BatchNorm2D & $-$    & $-$    & $16$     \\
Conv2D & $3\times3$    & $1\times1$    & $16$     \\
BatchNorm2D & $-$    & $-$    & $16$     \\
MaxPool2D & $2\times2$    & $2\times2$    & $16$     \\ \cdashline{1-4}
Conv2D & $3\times3$    & $1\times1$    & $64$     \\
BatchNorm2D & $-$    & $-$    & $64$     \\
Conv2D & $3\times3$    & $1\times1$    & $64$     \\
BatchNorm2D & $-$    & $-$    & $64$     \\
Conv2D & $3\times3$    & $1\times1$    & $64$     \\
BatchNorm2D & $-$    & $-$    & $64$     \\
MaxPool2D & $2\times2$    & $2\times2$    & $64$     \\
FC     &    $-$    &    $-$    & $128$       \\
FC     &    $-$    &    $-$    & $128$       \\
FC     &    $-$    &    $-$    & \# of classes       \\
\hline
\end{tabular}
\vspace{-10mm}
}

\end{wraptable}

\paragraph{\textit{Deep Learning Models.}}
For biomedical image classification tasks, we rely on a convolutional neural network \revise{containing five convolutional layers, two max-pooling layers, and three fully-connected (FC) layers with ReLU activation}. Table~\ref{tab:CNN_structure} provides the details of the network architecture.
The client owns the first part of the model, containing two convolutional layers and one max-pooling layer.
For the readmission prediction task on the eICU dataset, we apply the same model architecture as described in~\cite{yang2023manydg}. The client model utilizes separate encoders for mapping different types of events into a same-length embedding sequence, which is then processed by a Transformer encoder. The server model is a two-layer fully-connected network for making final predictions.
For the discharge prediction task with the VUMC dataset, we use a four-layer fully-connected network ($2808-64-32-32-1$) with ReLU activation,
where the model is split after the first layer.

\paragraph{\textit{Centralized/Distributed Learning Setting.}}
In the centralized learning scenario, we train the deep learning model on the entire training set for $50$ epochs using Adam optimizer with a batch size of $256$. By default, we use a learning rate of
$10^{-4}$ and a weight decay of $10^{-5}$. Specially, for the readmission prediction task on the eICU dataset, we set the learning rate to be $5\times10^{-4}$. For the discharge prediction task on the VUMC dataset, we use a weight decay of $10^{-4}$.
In the case of federated and split learning scenarios, we randomly partition the training dataset into $k$ disjoint subsets and assign them to each client. The parameters for the optimizers are set to be the same as in centralized learning.
Unless mentioned otherwise, we use a default number of clients $k=5$ in our experiments.

\begin{table}[t]
\centering
\caption{A comparison of final performance measured on the test dataset for centralized learning (CL), federated learning (FL), and split learning (SL) (reported with a $95\%$ confidence interval).}
\label{tab:final_result}
\resizebox{0.96\linewidth}{!}{
\begin{tabular}{c||cc|cc|cc|ccc|ccc}
\hline
\multirow{2}{*}{} & \multicolumn{2}{c|}{\textbf{PathMNIST}}                                                                                                    & \multicolumn{2}{c|}{\textbf{OrganAMNIST}}                                                                                                  & \multicolumn{2}{c|}{\textbf{BloodMNIST}}                                                                                                   & \multicolumn{3}{c|}{\textbf{eICU}}                                                                                                                                                                                          & \multicolumn{3}{c}{\textbf{VUMC}}                                                                                                                                                                                          \\ \cline{2-13} 
                  & \multicolumn{1}{c|}{\textbf{Accuracy}}                                         & \textbf{AUROC}                                              & \multicolumn{1}{c|}{\textbf{Accuracy}}                                         & \textbf{AUROC}                                              & \multicolumn{1}{c|}{\textbf{Accuracy}}                                         & \textbf{AUROC}                                              & \multicolumn{1}{c|}{\textbf{AUPRC}}                                            & \multicolumn{1}{c|}{\textbf{F1}}                                               & \textbf{Kappa}                                            & \multicolumn{1}{c|}{\textbf{AUPRC}}                                                     & \multicolumn{1}{c|}{\textbf{F1}}                                                        & \textbf{Kappa}                                                     \\ \hline \hline
\textbf{CL}       & \multicolumn{1}{c|}{\begin{tabular}[c]{@{}c@{}}0.8346\\$\pm$0.0088\end{tabular}} & \begin{tabular}[c]{@{}c@{}}0.9734\\$\pm$0.0071\end{tabular} & \multicolumn{1}{c|}{\begin{tabular}[c]{@{}c@{}}0.8773\\$\pm$0.0051\end{tabular}} & \begin{tabular}[c]{@{}c@{}}0.9901\\$\pm$0.0010\end{tabular} & \multicolumn{1}{c|}{\begin{tabular}[c]{@{}c@{}}0.9379\\$\pm$0.0022\end{tabular}} & \begin{tabular}[c]{@{}c@{}}0.9958\\$\pm$0.0005\end{tabular} & \multicolumn{1}{c|}{\begin{tabular}[c]{@{}c@{}}0.6389\\$\pm$0.0127\end{tabular}} & \multicolumn{1}{c|}{\begin{tabular}[c]{@{}c@{}}0.6083\\$\pm$0.0057\end{tabular}} & \begin{tabular}[c]{@{}c@{}}0.4735\\$\pm$0.0087\end{tabular} & \multicolumn{1}{c|}{\begin{tabular}[c]{@{}c@{}}0.7898\\$\pm$0.0058\end{tabular}} & \multicolumn{1}{c|}{\begin{tabular}[c]{@{}c@{}}0.7147\\$\pm$0.0037\end{tabular}} & \begin{tabular}[c]{@{}c@{}}0.6565\\$\pm$0.0042\end{tabular} \\ \hline
\textbf{FL}       & \multicolumn{1}{c|}{\begin{tabular}[c]{@{}c@{}}0.8225\\$\pm$0.0175\end{tabular}} & \begin{tabular}[c]{@{}c@{}}0.9690\\$\pm$0.0043\end{tabular} & \multicolumn{1}{c|}{\begin{tabular}[c]{@{}c@{}}0.8689\\$\pm$0.0065\end{tabular}} & \begin{tabular}[c]{@{}c@{}}0.9890\\$\pm$0.0010\end{tabular} & \multicolumn{1}{c|}{\begin{tabular}[c]{@{}c@{}}0.9165\\$\pm$0.0107\end{tabular}} & \begin{tabular}[c]{@{}c@{}}0.9932\\$\pm$0.0012\end{tabular} & \multicolumn{1}{c|}{\begin{tabular}[c]{@{}c@{}}0.6384\\$\pm$0.0122\end{tabular}} & \multicolumn{1}{c|}{\begin{tabular}[c]{@{}c@{}}0.5943\\$\pm$0.0101\end{tabular}} & \begin{tabular}[c]{@{}c@{}}0.4675\\$\pm$0.0098\end{tabular} & \multicolumn{1}{c|}{\begin{tabular}[c]{@{}c@{}}0.7884\\$\pm$0.0019\end{tabular}} & \multicolumn{1}{c|}{\begin{tabular}[c]{@{}c@{}}0.7122\\$\pm$0.0029\end{tabular}} & \begin{tabular}[c]{@{}c@{}}0.6547\\$\pm$0.0035\end{tabular} \\ \hline
\textbf{SL}       & \multicolumn{1}{c|}{\begin{tabular}[c]{@{}c@{}}0.8127\\$\pm$0.0129\end{tabular}} & \begin{tabular}[c]{@{}c@{}}0.9673\\$\pm$0.0069\end{tabular} & \multicolumn{1}{c|}{\begin{tabular}[c]{@{}c@{}}0.8538\\$\pm$0.0067\end{tabular}} & \begin{tabular}[c]{@{}c@{}}0.9864\\$\pm$0.0012\end{tabular} & \multicolumn{1}{c|}{\begin{tabular}[c]{@{}c@{}}0.9066\\$\pm$0.0096\end{tabular}} & \begin{tabular}[c]{@{}c@{}}0.9917\\$\pm$0.0013\end{tabular} & \multicolumn{1}{c|}{\begin{tabular}[c]{@{}c@{}}0.6331\\$\pm$0.0109\end{tabular}} & \multicolumn{1}{c|}{\begin{tabular}[c]{@{}c@{}}0.6000\\$\pm$0.0074\end{tabular}} & \begin{tabular}[c]{@{}c@{}}0.4693\\$\pm$0.0078\end{tabular} & \multicolumn{1}{c|}{\begin{tabular}[c]{@{}c@{}}0.7840\\$\pm$0.0028\end{tabular}} & \multicolumn{1}{c|}{\begin{tabular}[c]{@{}c@{}}0.7116\\$\pm$0.0026\end{tabular}} & \begin{tabular}[c]{@{}c@{}}0.6524\\$\pm$0.0033\end{tabular} \\ \hline
\end{tabular}
}
\end{table}

\begin{figure*}[t]
  \centering
  
  \begin{subfigure}{0.192\linewidth}
    \centering
    \includegraphics[width=\columnwidth]{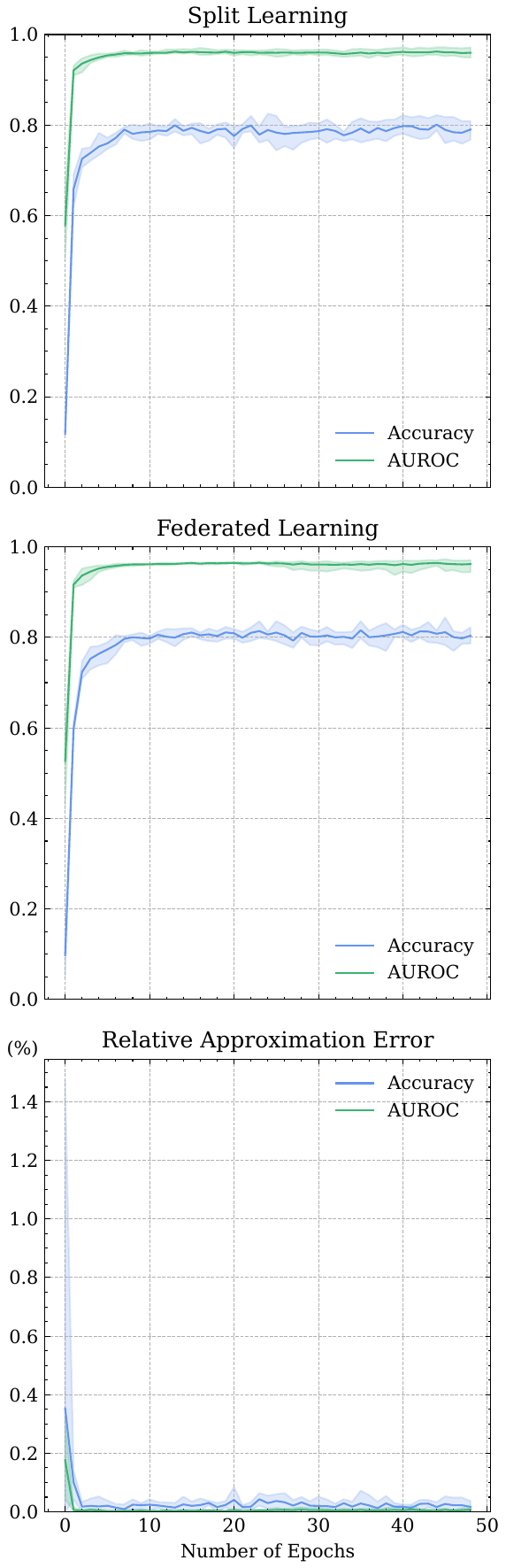}
    \caption{PathMNIST}
    \label{subfig:path}
  \end{subfigure}
  \hfill
  \begin{subfigure}{0.192\linewidth}
    \centering
    \includegraphics[width=\columnwidth]{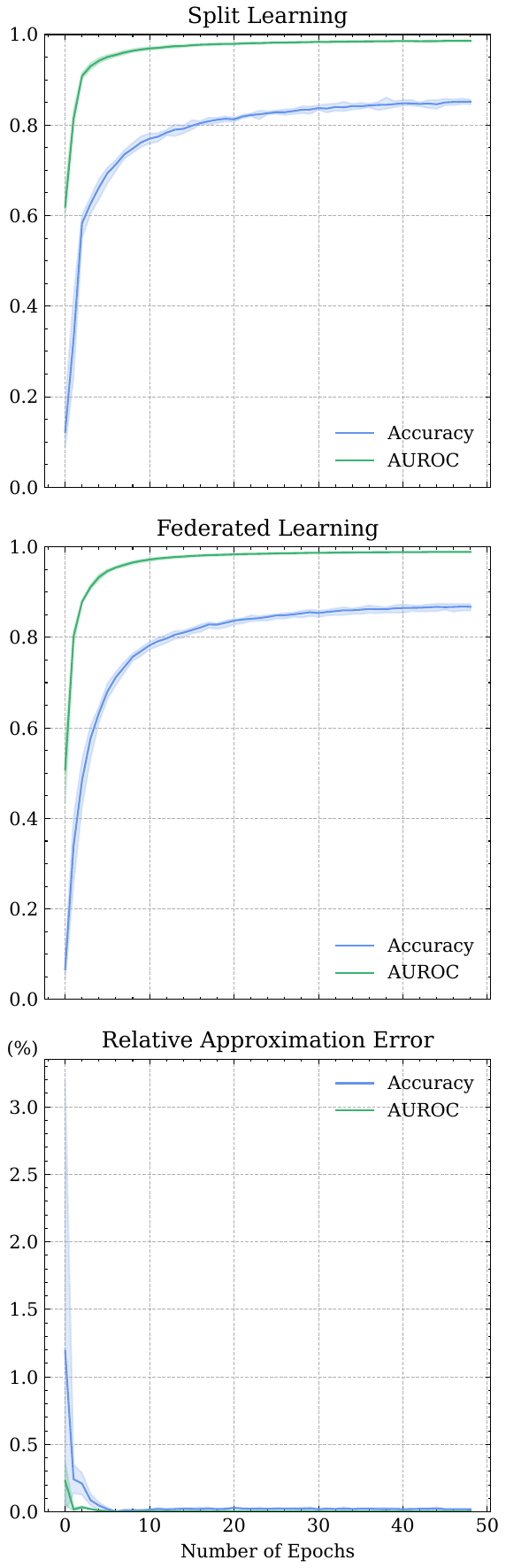}
    \caption{OrganAMNIST}
    \label{subfig:orga}
  \end{subfigure}
  \hfill
  \begin{subfigure}{0.192\linewidth}
    \centering
    \includegraphics[width=\columnwidth]{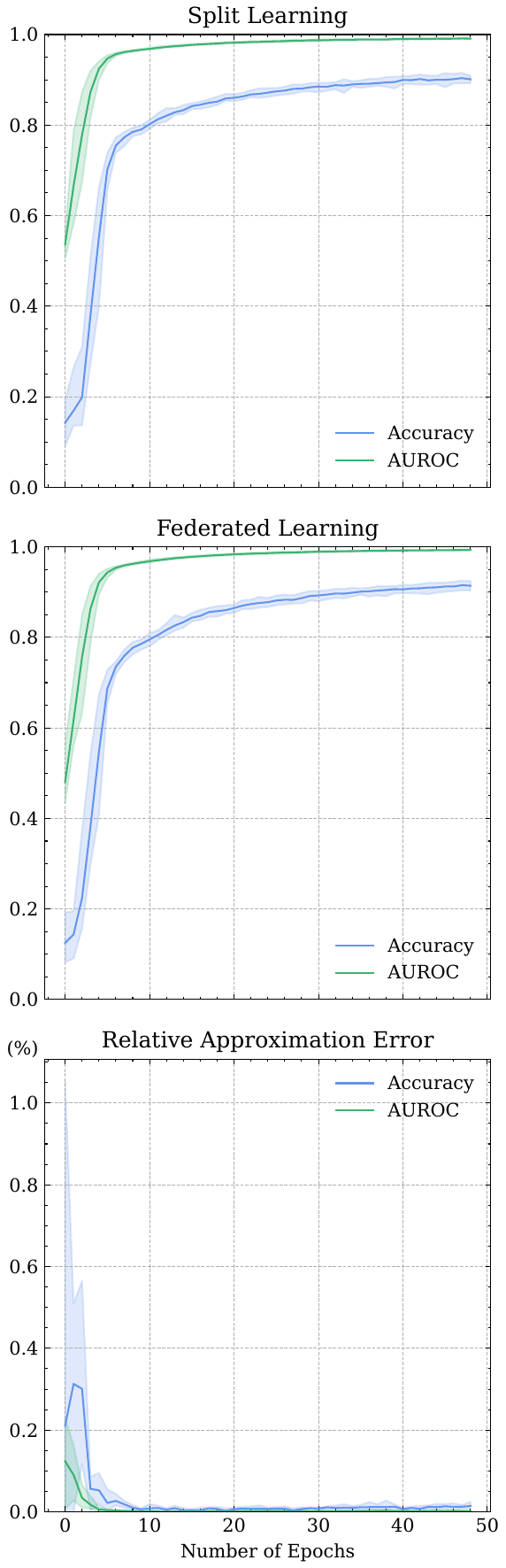}
    \caption{BloodMNIST}
    \label{subfig:bloo}
  \end{subfigure}
  \hfill
  \begin{subfigure}{0.192\linewidth}
    \centering
    \includegraphics[width=\columnwidth]{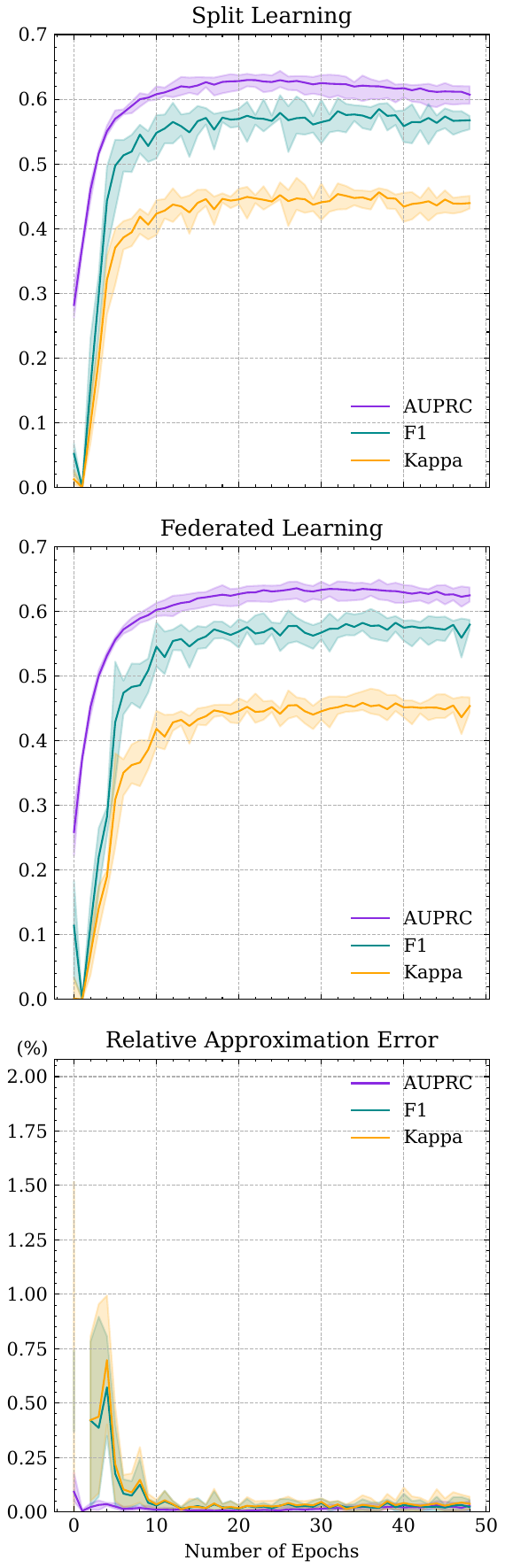}
    \caption{eICU}
    \label{subfig:eICU}
  \end{subfigure}
  \hfill
  \begin{subfigure}{0.192\linewidth}
    \centering
    \includegraphics[width=\columnwidth]{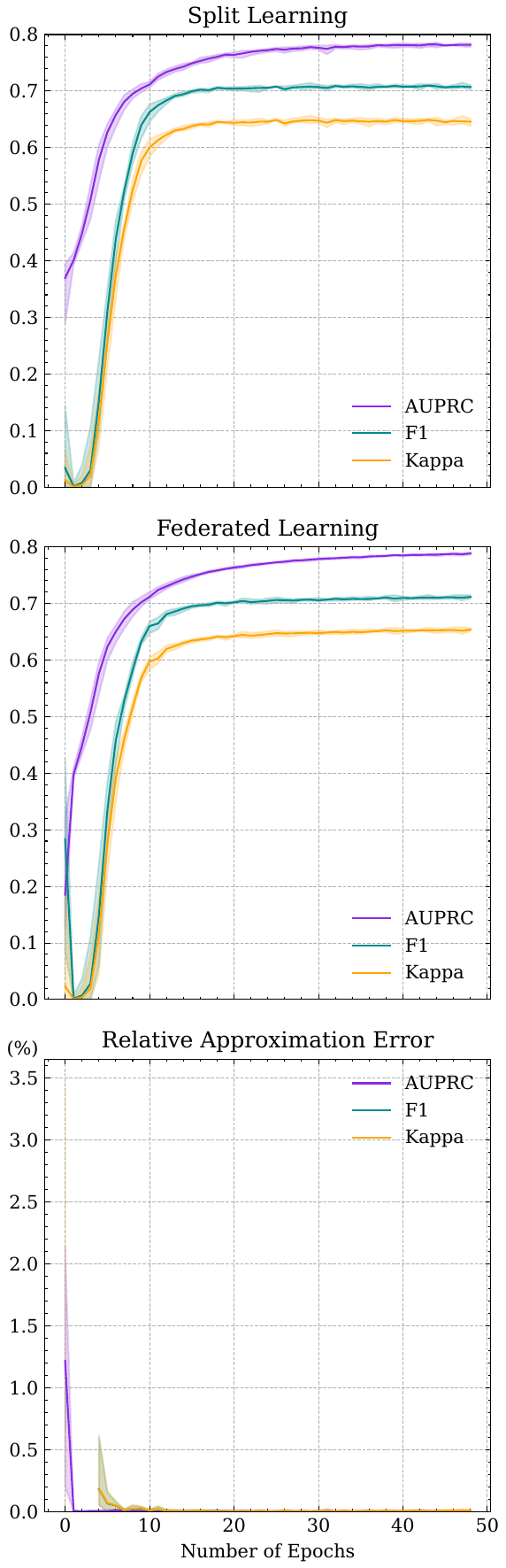}
    \caption{VUMC}
    \label{subfig:discharge}
  \end{subfigure}

  \vspace{1mm}
  \caption{Comparison of convergence by measuring the per-epoch performance on the test dataset for federated learning and split learning.}
  \vspace{1mm}

  \label{fig:convergence}
\end{figure*}
\begin{figure*}[t]
  \centering
  
  \begin{subfigure}{0.31\linewidth}
    \centering
    \includegraphics[width=0.475\columnwidth]{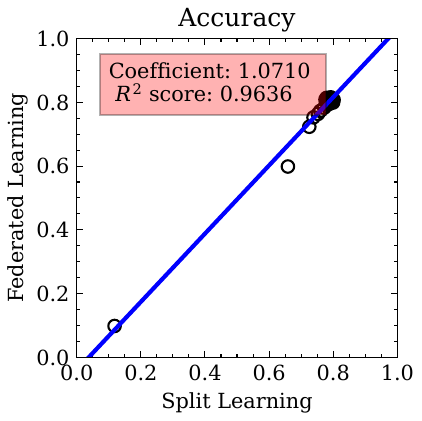}
    \includegraphics[width=0.475\columnwidth]{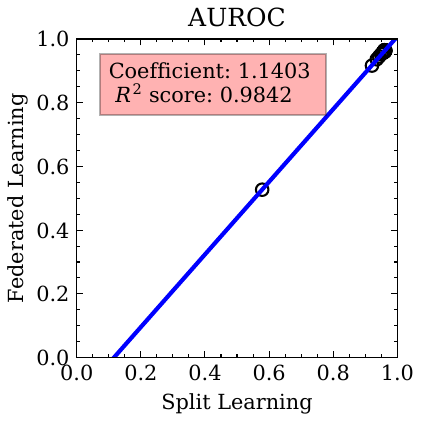}
    \caption{PathMNIST}
    \label{subfig:path_LR}
    \vspace{4mm}
  \end{subfigure}
  \hfill
  \begin{subfigure}{0.31\linewidth}
    \centering
    \includegraphics[width=0.475\columnwidth]{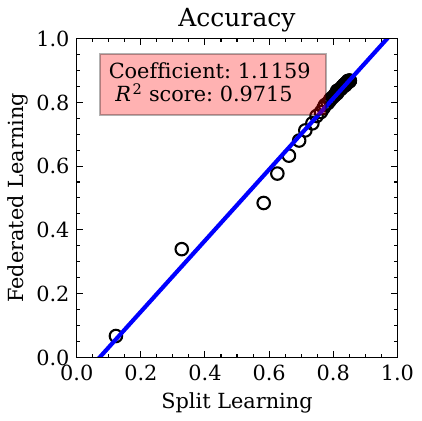}
    \includegraphics[width=0.475\columnwidth]{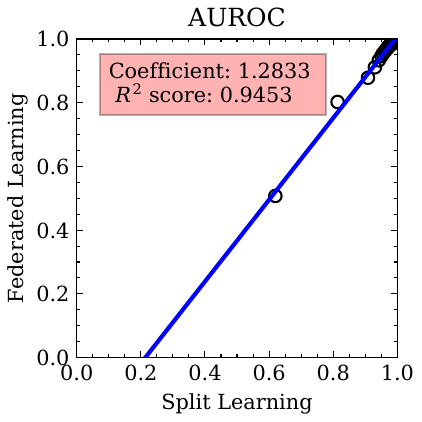}
    \caption{OrganAMNIST}
    \label{subfig:orga_LR}
    \vspace{4mm}
  \end{subfigure}
  \hfill
  \begin{subfigure}{0.31\linewidth}
    \centering
    \includegraphics[width=0.475\columnwidth]{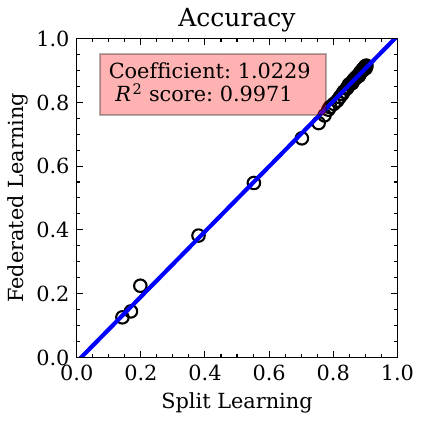}
    \includegraphics[width=0.475\columnwidth]{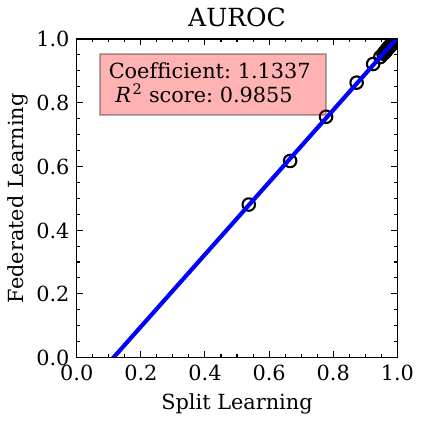}
    \caption{BloodMNIST}
    \label{subfig:bloo_LR}
    \vspace{4mm}
  \end{subfigure}

  \begin{subfigure}{0.47\linewidth}
    \centering
    \includegraphics[width=0.32\columnwidth]{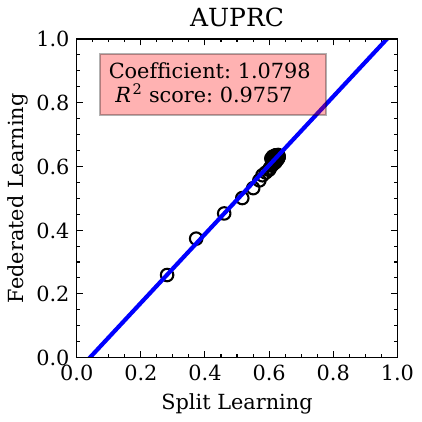}
    \includegraphics[width=0.32\columnwidth]{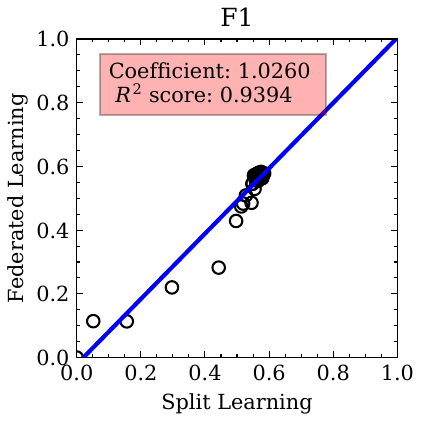}
    \includegraphics[width=0.32\columnwidth]{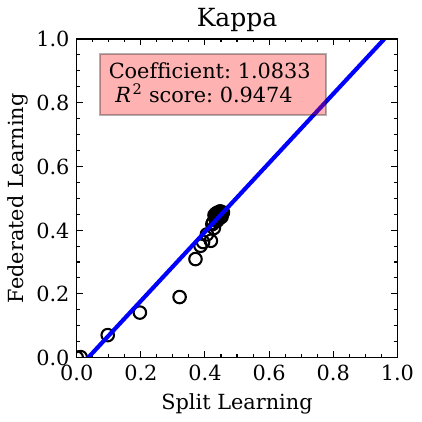}
    \caption{eICU}
    \label{subfig:eICU_LR}
  \end{subfigure}
  \hfill
  \begin{subfigure}{0.47\linewidth}
    \centering
    \includegraphics[width=0.32\columnwidth]{fig/LR/eICU_prauc_test_LR.pdf}
    \includegraphics[width=0.32\columnwidth]{fig/LR/eICU_f1_test_LR.pdf}
    \includegraphics[width=0.32\columnwidth]{fig/LR/eICU_cohen_test_LR.pdf}
    \caption{VUMC}
    \label{subfig:discharge_LR}
  \end{subfigure}

  \vspace{1mm}
  \caption{Linear regression results \revise{of the per-epoch performance of federated learning vs. split leaning} on $5$ datasets.}
  \label{fig:linear_regression}
  \vspace{1mm}

  \label{fig:out_of_dist}
\end{figure*}

\paragraph{\textit{Metrics.}}
We use the following metrics for the biomedical image classification tasks:
(1) \textit{Accuracy}: the average classification accuracy; and
(2) \textit{AUROC}: the average area under the receiver operating characteristic (ROC) curve\revise{s of each class against the rest}.
For the EHR prediction tasks, we report the following measures:
(1) \textit{AUPRC}: the average area under the precision-recall curve;
(2) \textit{F1}: the average F1 score; and
(3) \textit{Kappa}: Cohen's Kappa score, which measures the agreement between two raters on a classification task, defined as $\kappa=\frac{p_o - p_e}{1-p_e}$, where $p_o$ is the empirical probability of agreement on the label assigned to any sample (i.e., the observed agreement ratio), and $p_e$ is the expected agreement when both raters assign labels randomly. The Kappa ranges from $-1$ to $1$, with a larger value indicating a stronger agreement.
For each setting in our experiment, we train the neural network on the training dataset with five different random seeds and report its performance as measured on the hold-out testing dataset.

\begin{wrapfigure}{r}{0.25\linewidth}
  \centering
  \vspace{-4mm}
  \begin{subfigure}{1\linewidth}
    \centering
    \includegraphics[width=\columnwidth]{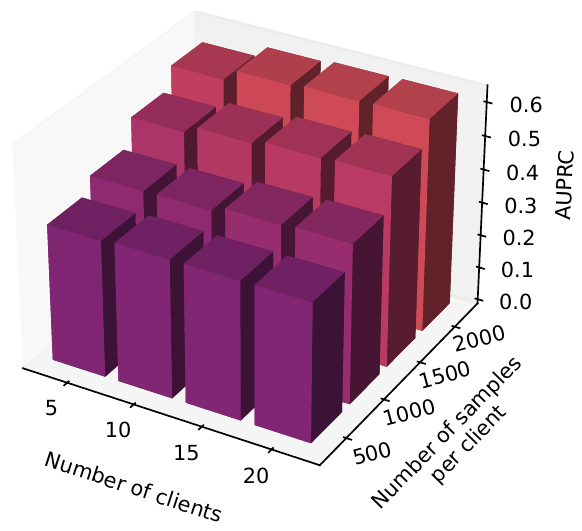}
    \caption{Federated learning}
    \label{subfig:impact_fl}
  \end{subfigure}
  \vspace{1mm}
  
  \begin{subfigure}{1\linewidth}
    \centering
    \includegraphics[width=\columnwidth]{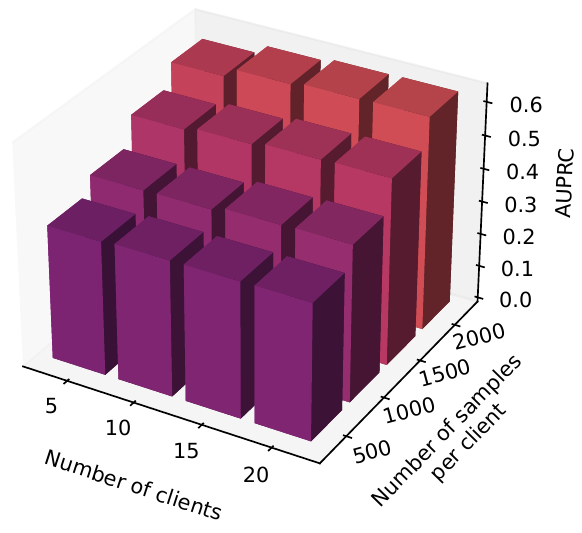}
    \caption{Split learning}
    \label{subfig:impact_sfl}
  \end{subfigure}
  \vspace{1mm}

  \begin{subfigure}{0.96\linewidth}
    \centering
    \includegraphics[width=\columnwidth]{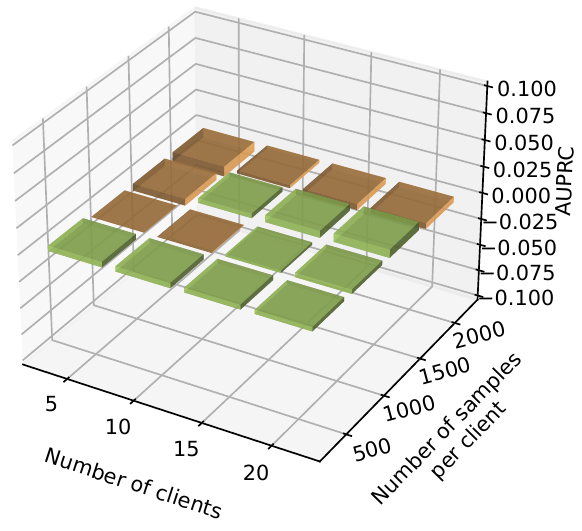}
    \caption{Difference}
    \label{subfig:impact_diff}
  \end{subfigure}
  \vspace{1mm}
  \caption{Sensitivity analysis on the VUMC dataset.}
  \label{fig:impact}
  \vspace{-6mm}
\end{wrapfigure}

\subsection*{\textit{Utility Analysis}}

\paragraph{\textit{Performance.}}
We conducted experiments using centralized learning, federated learning, and split learning under the same conditions for $50$ epochs. To eliminate the effects of overfitting in centralized learning, we \revise{assume early stopping} is employed and report the best performance measured on the test dataset. As shown in Table~\ref{tab:final_result}, the centralized learning consistently achieved the highest performance across all five datasets, which can be considered the upper bound of the performance for distributed learning. While federated learning outperformed split learning slightly on four datasets (excluding eICU), the difference in performance between them was almost negligible (usually $<1\%$), with margins of error being similar.

\paragraph{\textit{Convergence.}}
Figure~\ref{fig:convergence} compares the convergence of the global model between federated learning and split learning. The first and second rows plot the model performance measured on the test dataset after each epoch for split and federated learning, respectively. The third row plots the relative approximation error between federated and split learning, defined as $\delta=|\frac{v_{SL}-v_{FL}}{v_{FL}}|\times 100\%$, where $v_{SL}$ and $v_{FL}$ denote the performance of split and federated learning model respectively. We observe that except for the differences caused by the initialization at the beginning of training, both split and federated learning are able to converge at a similar rate. For some datasets (e.g., eICU and VUMC), split learning can converge even faster than federated learning. This observation was further verified by the linear regression analysis shown in Figure~\ref{fig:linear_regression}, which demonstrates a high correlation for all datasets (typically with estimated coefficients $< 1.2$ and R$^2$ $> 0.9$).

\paragraph{\textit{Scalibility.}}
To investigate the scalability of federated learning and split learning, we perform a sensitivity analysis on the VUMC dataset with different numbers of participating clients as well as different numbers of training data samples per client. As shown in Figure~\ref{fig:impact}, a similar pattern is observed for both federated learning and split learning, where the performance of the resulting model (measured by AUPRC) continues to improve as the number of clients and the number of samples per client increases. Moreover, the differences in performance between federated learning and split learning as shown in Figure~\ref{subfig:impact_diff} (green for positive and orange for negative values) are negligible. This indicates that federated learning and split learning are both scalable to more participating clients and a larger amount of training data.

\paragraph{\textit{Efficiency.}}
\begin{wraptable}{r}{0.5\linewidth}
\centering
\vspace{-1mm}
\caption{Comparison on client-side model efficiency.}
\vspace{-2mm}
\label{tab:flops}
\resizebox{\linewidth}{!}{
\begin{threeparttable}
\begin{tabular}{c||cc|cc}
\hline
\multirow{2}{*}{\textbf{Dataset}} & \multicolumn{2}{c|}{\textbf{Split Learning}}                & \multicolumn{2}{c}{\textbf{Federated Learning}}            \\ \cline{2-5} 
                                  & \multicolumn{1}{c|}{\textbf{\# of params}} & \textbf{FLOPs} & \multicolumn{1}{c|}{\textbf{\# of params}} & \textbf{FLOPs} \\ \hline \hline
PathMNIST                         & \multicolumn{1}{c|}{2,832}                  & 1,726,208        & \multicolumn{1}{c|}{235,225}                & 7,591,817        \\ \hline
OrganAMNIST                       & \multicolumn{1}{c|}{2,544}                  & 1,531,520        & \multicolumn{1}{c|}{235,195}                & 7,397,387        \\ \hline
BloodMNIST                        & \multicolumn{1}{c|}{2,832}                  & 1,726,208        & \multicolumn{1}{c|}{235,096}                & 7,591,688        \\ \hline
eICU                              & \multicolumn{1}{c|}{1,556,475}               & $-^*$              & \multicolumn{1}{c|}{1,558,556}               & $-^*$              \\ \hline
VUMC                              & \multicolumn{1}{c|}{179,776}                & 179,904         & \multicolumn{1}{c|}{186,049}                  & 186,369           \\ \hline
\end{tabular}
\begin{tablenotes}
  \small
  \item $^*$ Cannot be estimated since the input is heterogeneous clinical sequences with variable lengths and diverse features.
\end{tablenotes}
\end{threeparttable}
}
\vspace{-4mm}
\end{wraptable}

Table~\ref{tab:flops} compares the efficiency on the client's end in terms of model parameters and the estimated floating point operations (FLOPs) required for computing a model forward pass. Our results indicate that split learning consumes less memory and requires fewer computations on the client side compared to federated learning across all five datasets. Notably, split learning outperforms federated learning significantly in medical image recognition tasks, reducing memory usage by approximately $99\%$ and computational requirements by $77\%$. However, the benefits of split learning in EHR tasks are marginal, likely due to the complexity of prediction tasks, the specific model architecture, and the choice of cut layer position used in our experiments.

\subsection*{\textit{Privacy Analysis}}

Both federated learning and split learning mitigate the systemic privacy risks from traditional centralized learning by embedding a \textit{data minimization principle} in their design. Federated learning achieves this principle by aggregating information collected from multiple data records into a focused model update. By contrast, split learning limits the amount of collected information about each data record by only sharing its smashed data.

\begin{figure}[t]
  \centering
  \begin{subfigure}{0.3\linewidth}
    \centering
    \includegraphics[width=0.9\columnwidth]{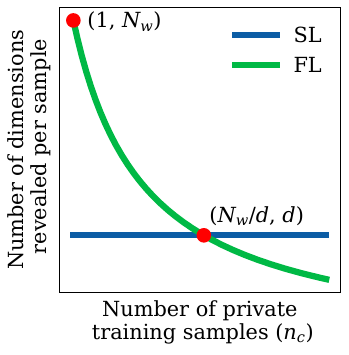}
    \caption{Privacy budget per sample measured by dimensionality}
    \label{subfig:privacy_per_sample}
  \end{subfigure}
  \hspace{1mm}
  \begin{subfigure}{0.4\linewidth}
    \centering
    \includegraphics[width=0.78\columnwidth]{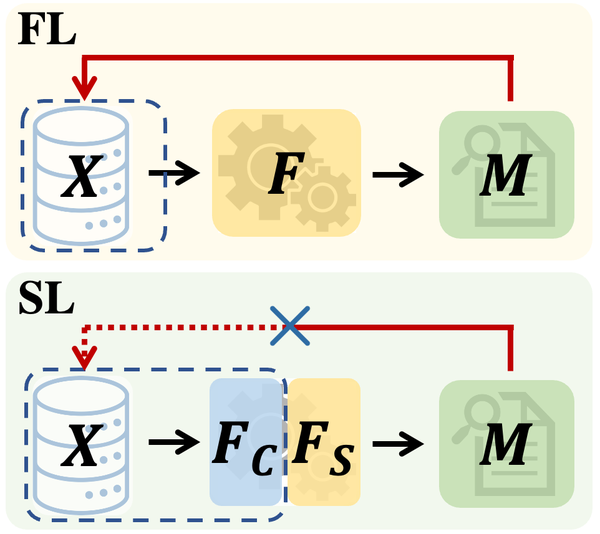}
    \caption{Privacy protection scheme: information within the blue dashed line are kept secret from the server}
    \label{subfig:privacy_scheme}
  \end{subfigure}

  \vspace{1mm}
  \caption{Conceptual comparison of privacy preservation in federated learning (FL) vs. split learning (SL).}
  \vspace{3mm}
\end{figure}

For the simplicity of analysis, herein we assume that the number of dimensions being revealed is proportional to the privacy risk, i.e., the amount of private information being leaked.
We note that this simplified privacy model is by no means a rigorous measure of data privacy, but it can serve as a baseline with room for refinement in the future.
More specifically, let us consider a client who contributes $n_c$ private data samples to participate in training. Suppose the federated learning model has $N_w$ the total number of parameters (i.e., $w \in \mathbb{R}^{N_w}$) and the cut layer size/smashed data dimension is $d$ (i.e., $h_{w^c}(x) \in \mathbb{R}^d$). Then during each training round, the average number of dimensions revealed to the server per each training sample is $\frac{N_w}{n_c}$ for federated learning and $d$ for split learning, as illustrated in Figure~\ref{subfig:privacy_per_sample}.
Since the model update is for a fixed number of dimensions (i.e., the same dimension as the model parameters), federated learning benefits from having a large size of local training data to average out the privacy risk.
By contrast, the amount of information revealed for each data sample in split learning is invariant to the local data size and only dependent on the client's model architecture (i.e., the cut layer size). Since deeper layers tend to produce more compact representations (i.e., smaller $d$), it is usually a trade-off between computational efficiency and privacy for the client model in split learning.
As such, federated learning appears to be more suitable for scenarios where every client possesses a sufficiently large quantity of data samples ($\geq \frac{N_w}{d}$) but discourages clients with fewer data from participating due to the higher privacy risk.
In our setting, the minimum reasonable data size for federated learning ($\frac{N_w}{d}$) is $102$, $2907$, and $12176$ for the biomedical image datasets, the VUMC dataset, and the eICU dataset, respectively. However, since modern deep neural networks are experiencing exponential growth in size, the minimum required data size can easily scale up to a number that is very difficult to achieve in practice. For example, if we train a ResNet-50 model~\cite{he2016deep} that contains over $23$ million trainable parameters to recognize medical images and choose to split after the adaptive pooling layer (\revise{corresponds to a cut layer size of $512$}), we would need at least over $44000$ data records from every site to justify the adoption of federated learning.
Such a requirement is oftentimes unattainable for healthcare applications, where a client with only a few patient records can also make a crucial contribution to the model.
For instance, specialty healthcare facilities (e.g., smaller oncology practices) may have much fewer patients compared to general hospitals, but they typically have a much more focused dataset in terms of certain diseases.
Therefore, split learning becomes more desirable in these circumstances as it is more versatile and can provide the same level of privacy benefits regardless of the size of the client's local data.

\paragraph{\textit{Risk of Inversion.}}
Compared to federated learning, split learning further reduces the risks of model/gradient inversion attack~\cite{fredrikson2015model,zhang2020secret,zhu2019deep,li2022auditing}
by \revise{restricting model access}
for both parties. As is depicted in Figure~\ref{subfig:privacy_scheme}, learning can be seen as computing a function $F$ on the client's private data $X$ to get result $M$.
In federated learning, although the raw data $X$ is not directly accessible to the server, the computation function $F$ and the result $M$ are both known to the server, which leaves the opportunity for the server to approximate an inverse function
to infer $X$ from $M$ and break patient privacy.
Differently, split learning separates the function $F$ (i.e., the model) into the client's part $F_C$ and the server's part $F_S$ and only the latter is revealed to the server. Thus, the server in split learning only has imperfect information about the computing process, which makes it quite challenging to compute the exact inverse function.

\section*{Discussion}

\paragraph{\textit{Design Trade-offs.}} There exist two major trade-offs in designing a deep learning model for split learning.
(1) \textit{Privacy-utility trade-off}: Split learning reflects the data minimization principle by allowing clients to extract and send only the relevant features of their local data to the server, while keeping the rest of the data private. The client model serves as a compressor which helps to reduce the amount of information that is transmitted over the network while preserving the most important information for the task at hand.
Using a small cut layer size further suppresses the amount of information in the smashed data, but may negatively affect the model performance. Thus, one of the main objectives of designing a split learning model is to choose a proper cut layer size to find a balance between keeping enough information to capture the important aspects of the data and reducing the amount of information to minimize the privacy risk of releasing the smashed data.
(2) \textit{Privacy-efficiency trade-off}: the information bottleneck theory~\cite{tishby2015deep} suggests that the output of deeper layers in a neural network contains more information about the task label and less information about the original input. As such, choosing to split at a deeper layer may result in less redundant information in the smashed data, thereby reducing the privacy risk. However, setting a shallow layer as the cut layer allows the client to have a more compact model and can thus give full scope to the advantages of split learning to reduce the memory consumption and computational burden on the client side.
Overall, these tuning knobs in split learning offer the user great flexibility in terms of configuring the privacy-utility-efficiency trade-off for specific tasks, whereas federated learning lacks such customizability.

\paragraph{\textit{Limitations.}} Despite the merits of split learning, there are several limitations we would like to highlight.
First, the current split learning framework only supports decentralized training of deep neural networks but does not support the training of traditional machine learning models, such as tree-based models.
Second, although split learning largely relieves the computational cost on the client side, the communication cost is increased compared to federated learning, as the client needs to communicate more frequently with the server for every mini-batch to perform gradient descent. To address this, recent studies propose to utilize an intermediate edge server~\cite{turina2021federated}, asynchronous training scheme~\cite{duan2022combined}, \revise{or an automated software framework~\cite{gharibi2022automated}} to reduce the communication overheads in split learning.
Additionally, service providers who do not wish to claim intellectual property over the developed model can share the server-side model with the clients after training to eliminate the communication overheads at inference time.
Third, the privacy benefits of split learning heavily rely on the inaccessibility of the client model and thus might be vulnerable under a stronger threat model. For instance, in the insider attack scenario where the server is able to collude with one of the clients, the privacy protection over other clients' data would be greatly diminished.

\paragraph{\textit{Future Directions.}}
There are several notable problems that should be considered as future research. First, how can we combine split learning with rigorous statistical privacy methods (e.g., differential privacy) or secure multiparty computation to achieve more rigorous privacy protection? Moreover, how can this be done with minimal sacrifice to model utility or significantly increasing the computational cost? Second, how can we design a more nuanced framework for privacy in split learning?  Specifically, we need to quantify the amount of leaked private information via smashed data and the potential privacy risks. Third, can we design a systematic framework to decide what the optimal cut layer size and location for a given neural network is to minimize the privacy risk while preserving most model utility?  Fourth, how can we realize split learning on heterogeneous data types and varying data distributions?

\subsection*{Conclusion}
A lack of sufficient data to cover the distribution of the general population is one of the major impediments to developing practical deep learning models for healthcare applications.
In this work, we \revise{introduced} split learning as a new distributed learning paradigm for enabling multi-institutional collaborative development of deep learning models across data silos without accessing raw patient data. Through both in-depth qualitative analysis as well as systematic quantitative experiments on five health datasets, we illustrated that split learning can achieve similar model utility as federated learning while providing better client-side model efficiency, lower risk of inversion, enhanced protection over the model, and more flexible privacy protection over clients' data. Our findings suggest that split learning is a promising alternative to federated learning for developing deep learning models without violating the privacy of the data contributors in many healthcare tasks.

\subparagraph{Acknowledgments}
This research was sponsored in part by grant U54HG012510, the NIH Bridge2AI Center. BM has been a paid consultant to TripleBlind AI, but for work unrelated to this investigation.

\makeatletter
\renewcommand{\@biblabel}[1]{\hfill #1.}
\makeatother

\bibliographystyle{vancouver}
\bibliography{amia}

\begin{thebibliography}{10}

\bibitem{esteva2017dermatologist}
Esteva A, Kuprel B, Novoa RA, Ko J, Swetter SM, Blau HM, et~al.
\newblock Dermatologist-level classification of skin cancer with deep neural
  networks.
\newblock Nature. 2017;542(7639):115-8.

\bibitem{wen2019desiderata}
Wen A, Fu S, Moon S, El~Wazir M, Rosenbaum A, Kaggal VC, et~al.
\newblock Desiderata for delivering NLP to accelerate healthcare AI advancement
  and a Mayo Clinic NLP-as-a-service implementation.
\newblock NPJ Digit Med. 2019;2(1):130.

\bibitem{landi2020deep}
Landi I, Glicksberg BS, Lee HC, Cherng S, Landi G, Danieletto M, et~al.
\newblock Deep representation learning of electronic health records to unlock
  patient stratification at scale.
\newblock NPJ Digit Med. 2020;3(1):96.

\bibitem{cohen2021problems}
Cohen JP, Cao T, Viviano JD, Huang CW, Fralick M, Ghassemi M, et~al.
\newblock Problems in the deployment of machine-learned models in health care.
\newblock CMAJ. 2021;193(35):E1391-4.

\bibitem{mcmahan2017communication}
McMahan B, Moore E, Ramage D, Hampson S, y~Arcas BA.
\newblock Communication-efficient learning of deep networks from decentralized
  data.
\newblock In: Artificial intelligence and statistics. PMLR; 2017. p. 1273-82.

\bibitem{rieke2020future}
Rieke N, Hancox J, Li W, Milletari F, Roth HR, Albarqouni S, et~al.
\newblock The future of digital health with federated learning.
\newblock NPJ Digit Med. 2020;3(1):119.

\bibitem{kaissis2020secure}
Kaissis GA, Makowski MR, R{\"u}ckert D, Braren RF.
\newblock Secure, privacy-preserving and federated machine learning in medical
  imaging.
\newblock Nature Machine Intelligence. 2020;2(6):305-11.

\bibitem{dayan2021federated}
Dayan I, Roth HR, Zhong A, Harouni A, Gentili A, Abidin AZ, et~al.
\newblock Federated learning for predicting clinical outcomes in patients with
  COVID-19.
\newblock Nat Med. 2021;27(10):1735-43.

\bibitem{abadi2016deep}
Abadi M, Chu A, Goodfellow I, McMahan HB, Mironov I, Talwar K, et~al.
\newblock Deep learning with differential privacy.
\newblock In: Proceedings of the 2016 ACM SIGSAC Conference on Computer and
  Communications Security; 2016. p. 308-18.

\bibitem{truex2020ldp}
Truex S, Liu L, Chow KH, Gursoy ME, Wei W.
\newblock LDP-Fed: Federated learning with local differential privacy.
\newblock In: Proceedings of the Third ACM International Workshop on Edge
  Systems, Analytics and Networking; 2020. p. 61-6.

\bibitem{constable2015privacy}
Constable SD, Tang Y, Wang S, Jiang X, Chapin S.
\newblock Privacy-preserving GWAS analysis on federated genomic datasets.
\newblock In: BMC Med Inform Decis Mak. vol.~15. BioMed Central; 2015. p. 1-9.

\bibitem{bonawitz2017practical}
Bonawitz K, Ivanov V, Kreuter B, Marcedone A, McMahan HB, Patel S, et~al.
\newblock Practical secure aggregation for privacy-preserving machine learning.
\newblock In: proceedings of the 2017 ACM SIGSAC Conference on Computer and
  Communications Security; 2017. p. 1175-91.

\bibitem{froelicher2021truly}
Froelicher D, Troncoso-Pastoriza JR, Raisaro JL, Cuendet MA, Sousa JS, Cho H,
  et~al.
\newblock Truly privacy-preserving federated analytics for precision medicine
  with multiparty homomorphic encryption.
\newblock Nat Commun. 2021;12(1):5910.

\bibitem{hardy2017private}
Hardy S, Henecka W, Ivey-Law H, Nock R, Patrini G, Smith G, et~al.
\newblock Private federated learning on vertically partitioned data via entity
  resolution and additively homomorphic encryption.
\newblock arXiv preprint arXiv:171110677. 2017.

\bibitem{gupta2018distributed}
Gupta O, Raskar R.
\newblock Distributed learning of deep neural network over multiple agents.
\newblock Journal of Network and Computer Applications. 2018;116:1-8.

\bibitem{vepakomma2018split}
Vepakomma P, Gupta O, Swedish T, Raskar R.
\newblock Split learning for health: Distributed deep learning without sharing
  raw patient data.
\newblock arXiv preprint arXiv:181200564. 2018.

\bibitem{kather2019predicting}
Kather JN, Krisam J, Charoentong P, Luedde T, Herpel E, Weis CA, et~al.
\newblock Predicting survival from colorectal cancer histology slides using
  deep learning: A retrospective multicenter study.
\newblock PLoS Med. 2019;16(1):e1002730.

\bibitem{bilic2019liver}
Bilic P, Christ PF, Vorontsov E, Chlebus G, Chen H, Dou Q, et~al.
\newblock The liver tumor segmentation benchmark (lits). arXiv.
\newblock arXiv preprint arXiv:190104056. 2019.

\bibitem{acevedo2020dataset}
Acevedo A, Merino A, Alf{\'e}rez S, Molina {\'A}, Bold{\'u} L, Rodellar J.
\newblock A dataset of microscopic peripheral blood cell images for development
  of automatic recognition systems.
\newblock Data in Brief. 2020;30.

\bibitem{pollard2018eicu}
Pollard TJ, Johnson AE, Raffa JD, Celi LA, Mark RG, Badawi O.
\newblock The eICU Collaborative Research Database, a freely available
  multi-center database for critical care research.
\newblock Sci Data. 2018;5(1):1-13.

\bibitem{zhang2021predicting}
Zhang X, Yan C, Malin BA, Patel MB, Chen Y.
\newblock Predicting next-day discharge via electronic health record access
  logs.
\newblock Journal of the American Medical Informatics Association.
  2021;28(12):2670-80.

\bibitem{thapa2022splitfed}
Thapa C, Arachchige PCM, Camtepe S, Sun L.
\newblock Splitfed: When federated learning meets split learning.
\newblock In: Proceedings of the AAAI Conference on Artificial Intelligence.
  vol.~36; 2022. p. 8485-93.

\bibitem{yang2023medmnist}
Yang J, Shi R, Wei D, Liu Z, Zhao L, Ke B, et~al.
\newblock MedMNIST v2-A large-scale lightweight benchmark for 2D and 3D
  biomedical image classification.
\newblock Scientific Data. 2023;10(1):41.

\bibitem{choi2020learning}
Choi E, Xu Z, Li Y, Dusenberry M, Flores G, Xue E, et~al.
\newblock Learning the graphical structure of electronic health records with
  graph convolutional transformer.
\newblock In: Proceedings of the AAAI Conference on Artificial Intelligence.
  vol.~34; 2020. p. 606-13.

\bibitem{yang2023manydg}
Yang C, Westover MB, Sun J.
\newblock ManyDG: Many-domain Generalization for Healthcare Applications.
\newblock In: Proceedings of the International Conference on Learning
  Representations; 2023. .

\bibitem{he2016deep}
He K, Zhang X, Ren S, Sun J.
\newblock Deep residual learning for image recognition.
\newblock In: Proceedings of the IEEE Conference on Computer Vision and Pattern
  Recognition; 2016. p. 770-8.

\bibitem{fredrikson2015model}
Fredrikson M, Jha S, Ristenpart T.
\newblock Model inversion attacks that exploit confidence information and basic
  countermeasures.
\newblock In: Proceedings of the 22nd ACM SIGSAC conference on computer and
  communications security; 2015. p. 1322-33.

\bibitem{zhang2020secret}
Zhang Y, Jia R, Pei H, Wang W, Li B, Song D.
\newblock The secret revealer: Generative model-inversion attacks against deep
  neural networks.
\newblock In: Proceedings of the IEEE/CVF Conference on Computer Vision and
  Pattern Recognition; 2020. p. 253-61.

\bibitem{zhu2019deep}
Zhu L, Liu Z, Han S.
\newblock Deep leakage from gradients.
\newblock Advances in neural information processing systems. 2019;32.

\bibitem{li2022auditing}
Li Z, Zhang J, Liu L, Liu J.
\newblock Auditing privacy defenses in federated learning via generative
  gradient leakage.
\newblock In: Proceedings of the IEEE/CVF Conference on Computer Vision and
  Pattern Recognition; 2022. p. 10132-42.

\bibitem{tishby2015deep}
Tishby N, Zaslavsky N.
\newblock Deep learning and the information bottleneck principle.
\newblock In: IEEE Information Theory Workshop. IEEE; 2015. p. 1-5.

\bibitem{turina2021federated}
Turina V, Zhang Z, Esposito F, Matta I.
\newblock Federated or split? a performance and privacy analysis of hybrid
  split and federated learning architectures.
\newblock In: Proceedings of the IEEE 14th International Conference on Cloud
  Computing. IEEE; 2021. p. 250-60.

\bibitem{duan2022combined}
Duan Q, Hu S, Deng R, Lu Z.
\newblock Combined federated and split learning in edge computing for
  ubiquitous intelligence in internet of things: State-of-the-art and future
  directions.
\newblock Sensors. 2022;22(16):5983.

\bibitem{gharibi2022automated}
Gharibi G, Patel R, Khan A, Gilkalaye BP, Vepakomma P, Raskar R, et~al.
\newblock An automated framework for distributed deep learning--a tool demo.
\newblock In: Proceedings of the IEEE 42nd International Conference on
  Distributed Computing Systems. IEEE; 2022. p. 1302-5.

\end{thebibliography}

\end{document}